\documentclass[a4paper,UKenglish,cleveref, autoref, thm-restate]{oasics-v2021}

\title{Automatic Modeling of Social Concepts Evoked by Art Images as Multimodal Frames}

\titlerunning{Automatic Modeling of Social Concepts} 

\author{Delfina Sol Martinez Pandiani\footnote{Corresponding author.}}{Department of Computer Science and Engineering (DISI), University of Bologna, Italy \and \url{https://www.unibo.it/sitoweb/delfinasol.martinez2/en} }{delfinasol.martinez2@unibo.it}{https://orcid.org/0000-0003-2392-6300}{}

\author{Valentina Presutti}{Department of Modern Languages, Literatures and Cultures (LILEC), University of Bologna, Italy}{valentina.presutti@unibo.it}{[http://orcid.org/0000-0002-9380-5160]}{}

\authorrunning{D.\,S.\,Martinez Pandiani and V. Presutti}

\Copyright{Delfina Sol Martinez Pandiani and Valentina Presutti} 

\ccsdesc[500]{Information systems~Content analysis and feature selection}
\ccsdesc[500]{Information systems~Ontologies}
\ccsdesc[500]{Information systems~Digital libraries and archives}
\ccsdesc[500]{Information systems~Multimedia information systems}

\keywords{Knowledge Engineering, Multisensory Data, Abstract Concepts, Image Understanding} 

\funding{This project has received funding from the European Union’s Horizon 2020 research and innovation programme under grant agreement No 101004746.}

\category{} 

\relatedversion{} 

\nolinenumbers 

\EventEditors{}
\EventNoEds{2}
\EventLongTitle{First International Workshop on Multisensory Data and Knowledge at the 3rd Conference on Language, Data and Knowledge (2021)}
\EventShortTitle{MDK 2021}
\EventAcronym{MDK}
\EventYear{2021}
\EventDate{September 1--3, 2021}
\EventLocation{Zaragoza, Spain}
\EventLogo{}
\SeriesVolume{}
\ArticleNo{}

\begin{document}
\maketitle

\begin{abstract}
Social concepts referring to non-physical objects--such as \emph{revolution}, \emph{violence}, or \emph{friendship}--are powerful tools to describe, index, and query the content of visual data, including ever-growing collections of art images from the Cultural Heritage (CH) field. While much progress has been made 
towards complete image understanding in computer vision, automatic detection of social concepts evoked by images is still a challenge. This is partly due to the well-known semantic gap problem, worsened for social concepts given their lack of unique physical features, and reliance on more unspecific features than concrete concepts. In this paper, we propose the translation of recent cognitive theories about social concept representation into a software approach to represent them as multimodal frames, by integrating multisensory data. Our method focuses on the extraction, analysis, and integration of multimodal features from visual art material tagged with the concepts of interest. We define a conceptual model and present a novel ontology for formally representing social concepts as multimodal frames. Taking the Tate Gallery’s collection as an empirical basis, we experiment our method on a corpus of art images to provide a proof of concept of its potential. 
We discuss further directions of research, and provide all software, data sources, and results.
\end{abstract}

\section{Introduction and Problem Statement}
\label{intro}

Rapidly expanding image collections, including ever-growing art catalogues from the Cultural Heritage (CH) field, have attracted significant research efforts for automatic retrieval and management of visual data. Social concepts--such as \emph{horror} or \emph{consumerism}--are a powerful resource for performing these tasks, especially of art images. This is because visual forms such as paintings and photographs are thought to illustrate, and circulate, concepts both by providing links to depicted objects through raw features--such as lines, color, shape, and size-- as well as through what Barthes called an image's {\lq connotation\rq}: a second layer of meaning made from culturally coded elements \cite{barthes1980camera}. For example, seeing Artemisia Gentileschi’s \emph{Judith Beheading Holofernes} (1620 ca.), a human observer is able to detect objects such as a sword, but a comprehensive understanding of the painting would generally include a social concept such as \emph{violence}. 

In this work, we aim to automatically model social concepts as multimodal frames based on features of art images that evoke them. We are interested in \emph{social concepts} in the ontological sense expounded by \cite{masolo2004social}--immaterial products whose conventional constitution involves a network of relations among social agents--which refer to \emph{non-physical} objects. The concepts of interest overlap with those which the field of cognitive neuroscience has labeled as {\lq abstract\rq} concepts, complex situations that do not possess a single and perceptually bounded object as referent \cite{borghi2018varieties}. For the rest of this paper, the term {\lq social concepts\rq} is used to allude specifically to this category of social concepts lacking unique, clearly perceivable referents.

The automatic modeling of social concepts based on perceptual features of art images could lead to breakthroughs in a wide range of applications, including Web image search, online picture-sharing communities, and scientific and academic endeavors, by serving as input to models that improve multimedia querying of digital libraries. Furthermore, it could improve automatic detection of the potential of images to communicate social concepts, a useful resource for CH institutions to enhance their visual collections’ documentation, descriptions, and mediation. Additionally, it could aid public institutions in building narratives about their visual objects, companies of the creative sector in exploiting their already existing catalogues, and companies building products or services related to specific social concepts.

While much progress has been made towards the general goal of complete image understanding in computer vision, including outstanding performance for tasks like image classification, object detection, and image generation \cite{bagi2020deep}, image understanding methods are not yet successful at detecting social concepts. This is partly because a main focus of computer vision has been image segmentation, while these concepts lack unique and distinctive physical features, such as defined form, color and texture. Instead, they tend to be more schematic in nature, with a critical parasitic relation between sensory-perceptual and distributional linguistic data given their \emph{acquired embodiment} \cite{davis2021building}. While there are multiple works focused on integrating knowledge in image understanding \cite{aditya2019integrating}, no frameworks have been developed to automatically model social concepts based on extraction and integration of sensory-perceptual data, such as pervasive visual features of images which evoke them, with distributional linguistic patterns of social concept usage. 

As such, our ultimate goal is to translate recent cognitive theories about concept representation into a software architecture that can automatically model social concepts based on multimodal features.
We hypothesize that a formal representation of social concepts as multimodal frames can be automatically produced with a pipeline combining knowledge engineering, computer vision, and computational linguistics methods. In this paper, we introduce our approach to generate these representations through the extraction and integration of features from images to Knowledge Graphs (KG). Our approach focuses on the extraction, analysis, and integration of multimodal features (including depicted concrete objects, depicted actions, and color features) from images tagged with social concepts. Taking the Tate Gallery collection as an empirical basis, we present a study of extraction and integration of multimodal data. The contributions of this paper can be summarised as follows:
\begin{itemize}
\item We define a conceptual model and present a novel, pattern-based ontology for social concept representation, which allows for semantic characterization of multimodal features.
\item We propose a novel approach for the extraction and integration of multimodal features of images that evoke certain social concepts.
\item We implement the proposed method on an corpus of art images from a well-known catalogue of art images (the Tate Gallery collection).
\item We discuss the results and provide all software, data sources, and results to allow the reproducibility of our experiment.
\end{itemize}

After discussing related work in Section 2, Section 3 describes the Tate Gallery Collection’s dataset, which was used as our experiments' input source. Section 4 illustrates our approach and its implementation. Section 5 reports our experiments and results, while Section 6 focuses on the discussion of the results and their limitations. We conclude with further directions in Section 7.

\section{Related Work}
\label{related}
Social concepts, in the scope of this work, refer to the immaterial products defined in \cite{masolo2004social}, where social concepts are ‘reified’ so as to be able to predicate on them, and concept definitions are explicitly introduced as descriptions that deal with the social, relational, and contextual nature of social concepts. Cognitive neuroscience has labeled such concepts {\lq abstract\rq}: with content more variable both within and across individuals, lacking a single and perceptually bounded object as referent, and classifiable in subtypes such as numerical, emotional, aesthetic, ethical, and moral concepts \cite{borghi2018varieties}. These concepts have previously been characterized as linguistic frames \cite{petersen2015representation}, a view mirrored in lexical resources such as FrameNet.\footnote{\url{https://framenet.icsi.berkeley.edu/fndrupal/}} Recent cognitive theories pose that, in addition to linguistic patterns, knowledge of abstract words is acquired through \emph{acquired embodiment}, whereby they become indirectly associated with the perceptual features of co-occurring concrete words \cite{hoffman2018concepts}.

In the CH field, social concepts have been recognised as a resource to manage visual collections, through the ample use of controlled thesauri such as the Getty vocabularies\footnote{\url{https://www.getty.edu/research/tools/vocabularies/}} and classifications systems like Iconclass\footnote{\url{http://iconclass.org/}} containing ready-made social concepts to be associated as subject matters of visual materials \cite{rafferty2017indexing}. Given how challenging it can be to computationally interpret culturally coded elements of visual materials, multidisciplinary frameworks such as \emph{distant viewing} \cite{arnold2019distant}, are calling for the construction of explicit code systems to automatically extract and interpret semantic elements of visual materials via techniques from exploratory data analysis.

Automatic detection of social concepts, even in natural images, is scarce, partly because almost all trainable classes in popular datasets used for image classification, such as ImageNet\footnote{\url{http://www.image-net.org/update-mar-11-2021.php}} and Google Open Images,\footnote{\url{https://opensource.google/projects/open-images-dataset}} refer to \emph{concrete} objects. A few works have attempted to train convolutional neural networks to detect more abstract classes in natural images: in \cite{ahres2016abstract}, one was trained to learn and detect abstract concepts and emotions in images using Flickr images and their tags, and in \cite{stabinger2017evaluation}, another was trained to automatically classify images of chessboards by the compositional elements \emph{identity} and \emph{symmetry}.
Most related efforts have focused on classifying images by evoked emotions through visual sentiment analysis \cite{chen2014deepsentibank}. The multidisciplinary research field of affective computing sees the usage of multimodal big data in the form of text, audio, visual and physiological signals as an avenue for more robust systems \cite{shoumy2020multimodal}. Most recently, Google Arts \& Culture and the University of California, Berkeley, released an interactive map of twenty-five emotions evoked by artworks based on the results of a user-based study.\footnote{\url{https://artsexperiments.withgoogle.com/art-emotions-map/}} 

Computer-vision based tools have greatly benefited the CH sector by improving access to implicit cultural knowledge contained in CH images, with novel methods and technologies being used in the last few years to perform image classification, object detection, caption generation, and more in art images \cite{diaz2020comparative}. Various feature extraction techniques and classification algorithms are available for 2D object recognition, including ones that use domain transfer to perform object recognition in paintings and art images \cite{crowley2016art}. Furthermore, there is an increasing number of techniques to train neural networks specifically to process art images, such as using style transfer to generate novel art image datasets from non artistic images, which are then used to finetune image processing algorithms (eg., \cite{kadish2021improving}).

Influenced by the idea that a cognitive understanding of our visual world requires that we complement computers' ability to detect objects with abilities to understand relationships within a scene, graphs are increasingly being employed to structure image information, with a growing number of datasets of high quality image \emph{scene graphs}, such as Visual Genome \cite{krishna2017visual}, regarding images' depicted objects, their attributes, and their relationships as first-class citizens of the annotation space. 
Additionally, because associating a meaning with complex scenes may require an explicit and symbolic representation of the domain of knowledge, ontologies have been shown to be a powerful framework to address image analysis and interpretation \cite{aditya2019integrating}, allowing queries to be formulated in terms of concepts and their relationships.

\section{Input Source: The Tate Gallery Collection}
\label{input}

As an empirical basis for our study, we use the Tate Gallery's collection metadata of 70,000 artworks made available as a Github repository.\footnote{ \url{https://github.com/tategallery/collection}} The Tate is an institution that houses the United Kingdom's national collection of British art, and international modern and contemporary art. Most of the collection is from the 1800s, and a considerable part of it is from after 1960. In 2013, the institution made their collection metadata available for about 70,000 artworks that it owns or jointly owns with the National Galleries of Scotland, through a Github repository. While the repository is no longer actively maintained, the Tate keeps it available as a useful tool for researchers and developers, and looks positively on creative remixing, visualization, and analysis of their collection metadata.\footnote{See Eric Drass' "Tate Explorer" at \url{http://shardcore.org/tatedata/} and Florian Kräutli's "The Tate Collection on Github" at \url{http://research.kraeutli.com/index.php/2013/11/the-tate-collection-on-github/}} 
The dataset contains complete records of most artists and artworks in the collection. It also includes image and thumbnail urls, but it does not directly provide images, which still need to be accessed online. The dataset can be accessed in two ways: either through CSV files containing information about the artists and artworks, or in a series of thousands of text files containing all the records in JSON format. The JSON data is much richer than the CSV, storing a list of subjects associated with the record organised in a subject taxonomy.

It was precisely because of its subject taxonomy that we selected the Tate Gallery Collection as a first dataset to test our approach; the rich taxonomy includes both concrete concepts and social concepts referring to non-physical objects (“vacuum cleaner,” “shoe”, “consumerism”, “horror”) as subject tags. As documented in an issue of their Github repository,\footnote{\url{ https://github.com/tategallery/collection/issues/27}} the Tate’s subject taxonomy is a “bespoke taxonomy”, originally developed alongside the digitisation of Tate's collection as a means of enabling visitors to search artworks via subject. The design of the hierarchical structure and initial tagging of the bulk of the collection was expert-led, carried out by indexers with art history backgrounds and with the support of Tate's curatorial team, and in consultation with pre-existing systems such as Iconclass and COLLAGE--the now unavailable public-access system for the Guildhall Library and Art Gallery.

\section{Approach}
\label{approach}

\begin{figure}[h]
    \centering
    \includegraphics[width=14cm]{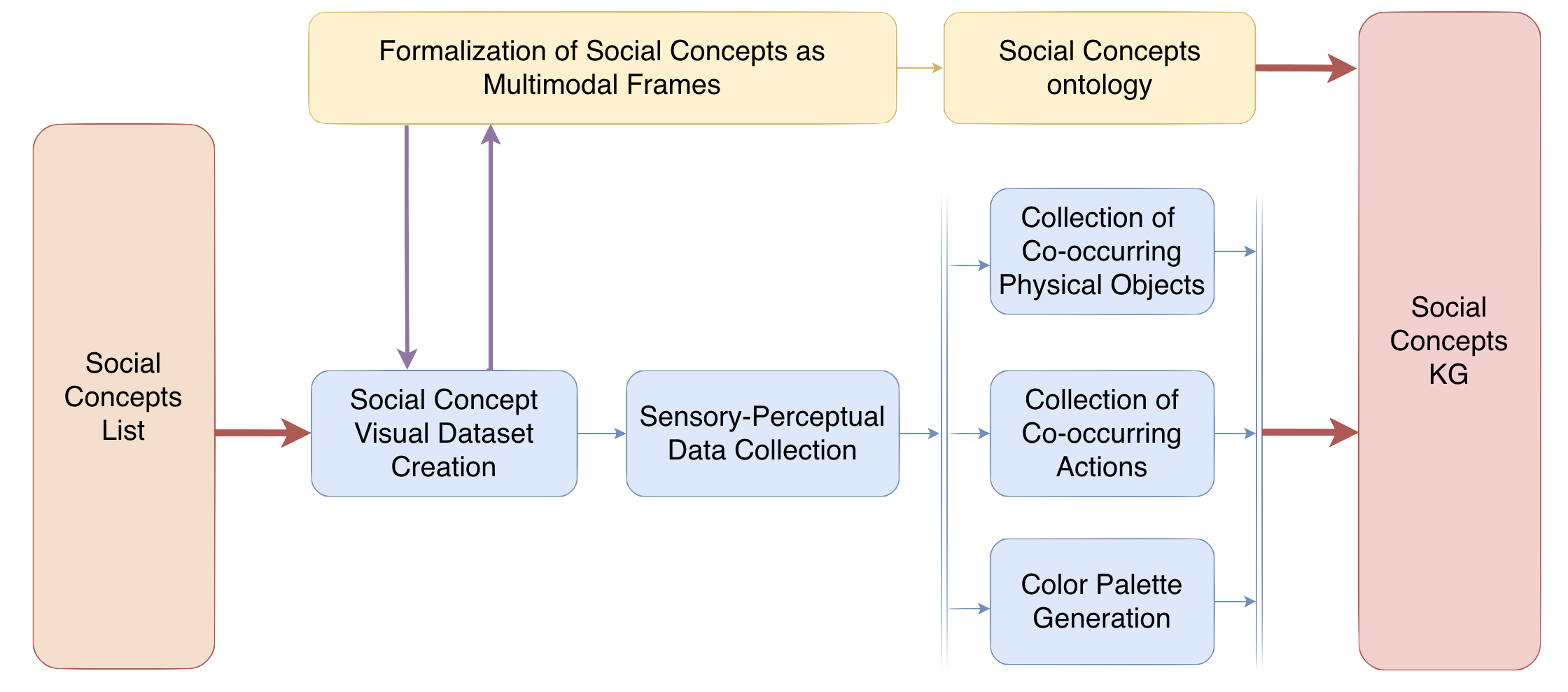}
    \caption{The pipeline under development aims to populate a large scale, ontology-based Social Concepts Knowledge Graph that describes social concepts with multimodal frames.}
    \label{fig:pipeline}
\end{figure}

Our approach is based on the idea that a social concept is a complex object whose definition can be formalized as a \emph{description} \cite{masolo2004social}. Specifically, our assumption is that a social concept can be described with a multimodal frame, whose meaning arises from an integration of linguistic and sensory-perceptual features of content, such as images, that evoke that social concept. For example, the meaning of the social concept ``death'' may be described via a multimodal frame integrating properties of linguistically co-occurring terms (e.g., ``coffin'', ``burial'', ``accident'') and sensory-perceptual properties of images depicting scenes that evoke that social concept (e.g., the color black as a sensory-perceptual feature of a funeral scene, which evokes the social concept ``death'') \cite{davis2021building}. We developed a framework (summarized in Figure \ref{fig:pipeline}) to integrate multimodal features related to social concepts into a scalable ontology-based Knowledge Graph (KG).

\noindent \textbf{Social Concepts Candidate List Creation.} Determining a starting list of candidate concepts that can be reliably classified as referring to non-physical objects is necessary to begin this research. In our approach, the initial list is based on the conceptual taxonomy already in use by the Tate Collection to tag the content of visual artworks, which explicitly includes categories such as ``universal concepts'' and ``social comment''.

\medskip

\noindent \textbf{Social Concept Definition by Multimodal Frames.} Our approach is based on the idea that a social concept can be defined in a multimodal frame which describes and integrates complex linguistic and sensory-perceptual features. In order to represent this model formally, we designed the Multimodal Descriptions of Social Concepts (MUSCO) ontology, based on the Descriptions and Situations (DnS) ontology \cite{gangemi2003understanding}, which supports a first-order manipulation of theories and models. DnS was chosen as a core design pattern because it allows for the modeling of non-physical objects, such as social concepts, whose intended meaning results from statements, i.e. they arise in combination with other entities. Specifically, DnS axioms capture the notion of \emph{situation} as a unitarian entity out of a State of Affairs (SoA), that is constituted by the entities and the relations among them, and a \emph{description} as an entity that partly represents a (possibly formalized) theory that can be conceived by an agent. 

\begin{figure}[h]
    \centering
    \includegraphics[width=14cm]{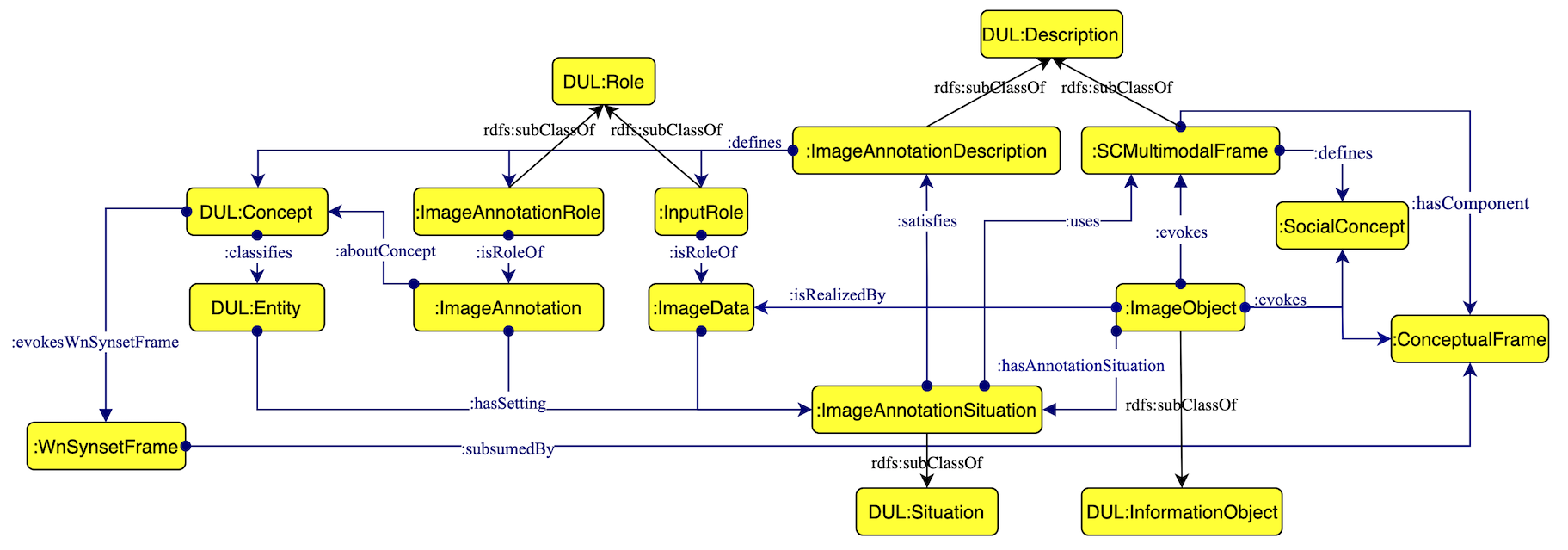}
    \caption{The MUSCO Ontology is aligned to and reuses patterns from DOLCE+DnS Ultralite (DUL) foundational ontology \cite{dul111} in order to represent and give meaning to all data created during an image annotation process. All un-prefixed classes belong to the namespace of the MUSCO ontology.}
    \label{fig:T_Box_0}
\end{figure}

Influenced by the work in \cite{vacura2008describing}, in the MUSCO ontology we consider that the image annotation process is a situation (i.e. a context reified in the class \texttt{ImageAnnotationSituation}) that needs to be described via a \texttt{ImageAnnotationDescription}, and that represents the state of affairs of all related, annotated data: actual multimedia data as well as metadata (see Figure \ref{fig:T_Box_0}).

\begin{figure}[h]
    \centering
    \includegraphics[width=14cm]{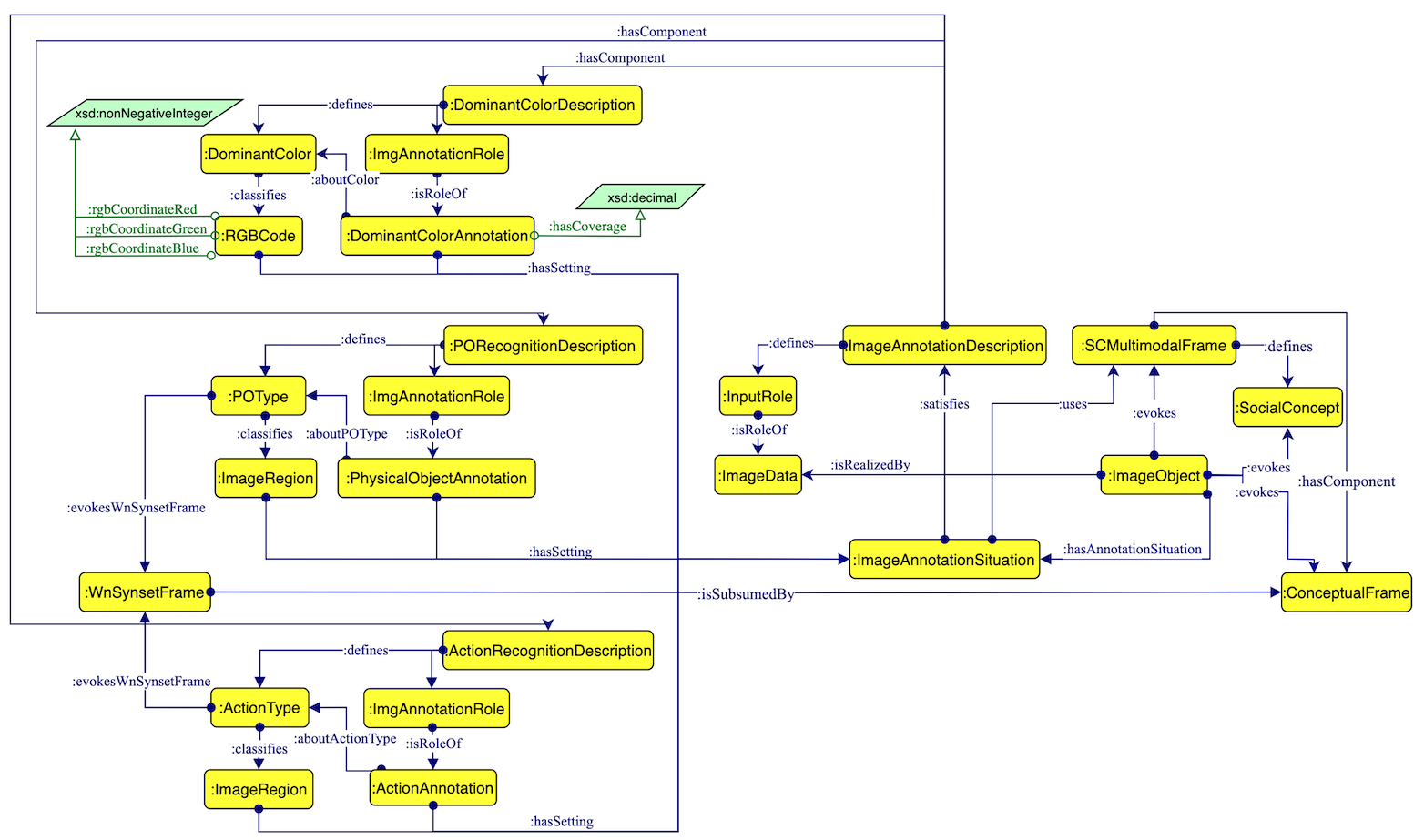}
    \caption{The MUSCO Ontology's reuse of the DnS pattern is modular: defining a general description for the image annotation situation, composed of simpler, more specific descriptions which give meaning to specific annotation structures and results. All classes in the figure belong to the namespace of the MUSCO ontology defined in this work.}
    \label{fig:T_Box_1}
\end{figure}

\noindent MUSCO's architecture is deliberately modular, such that an \texttt{ImageAnnotationDescription} can be modularized into sub-descriptions that capture complex structures to be annotated. At this stage, we have identified three complex structures to be annotated in art images evoking certain social concepts: dominant colors, depicted physical objects, and depicted actions. As such, in the MUSCO ontology we define the general description \texttt{ImageAnnotationDescription}, which is satisfied by the general \texttt{ImageAnnotationSituation}, and which is composed by three more specific descriptions (\texttt{DominantColorDescription}, \texttt{PORecognitionDescription}, and \texttt{ActionRecognitionDescription}) that define concepts and give meaning to data extracted in the context of each of the complex structures (see Figure \ref{fig:T_Box_1}). Finally, the MUSCO Ontology defines the description class \texttt{SCMultiModalFrame}, which (a) defines a \texttt{SocialConcept}, (b) is used by a \texttt{ImageAnnotationSituation}, and (c) is evoked by an \texttt{ImageObject}. The ontology also already allows for the conjunct expression of data coming from linguistic and lexical resources through classes such as \texttt{WnSynsetFrame} and \texttt{ConceptualFrame}. Even though the current version of the MUSCO ontology is still under a process of revision and editing, in its current state, it already includes the entities, resources, and relationships necessary to integrate the experimental outputs presented in this paper.

\begin{figure}[!h]
  \centering
  \includegraphics[width=1\textwidth]{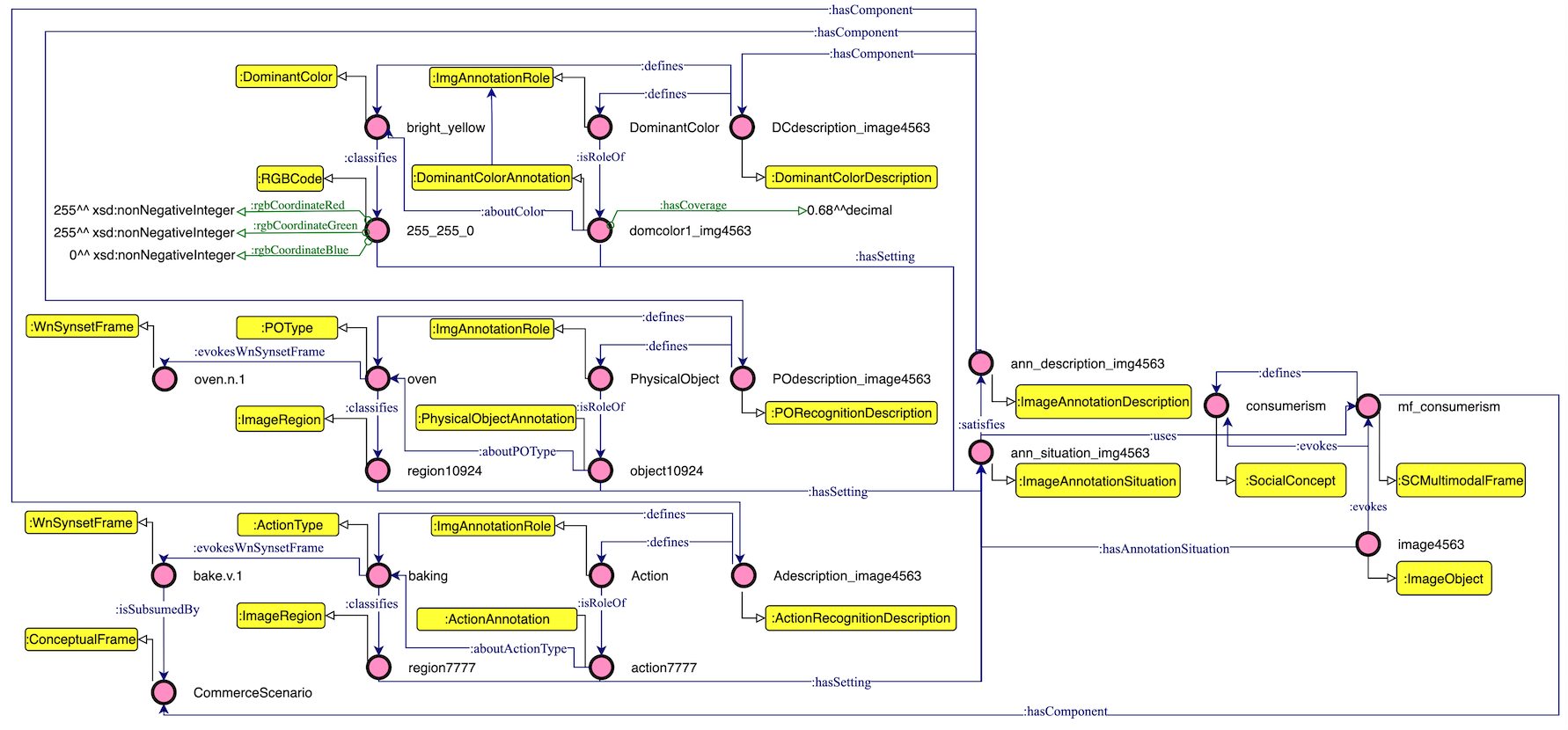}
    \caption{Example use of the MUSCO ontology to formalize multimodal features extracted from one image (dominant color \emph{bright yellow}, depiction of the physical object \emph{oven}, and depiction of the action \emph{baking}), the image's evocation of the social concept ``consumerism'', and the concept's description as a multimodal frame. All arrows with white arrowheads stand for the relation \texttt{rdf:type}.}
\label{fig:A_Box}
\end{figure}

\medskip
\noindent \textbf{Social Concept Visual Dataset Creation.} For each social concept, a corpus of (art) images that have previously been explicitly tagged with that social concept is created. This corpus is produced by performing surveys of existing image catalogues which include that social concept in their tagging scheme.

\medskip

\noindent \textbf{Sensory-Perceptual Data Collection.} Our approach is based on the idea that sensory-perceptual features of social concepts can be extracted from images that evoke those social concepts. For each social concept, based on its corpus of images, we extract a set of features including: labels for concrete objects, actions depicted, and dominant color palettes. The specific technique for extracting concrete objects labels from the images depends on the previously provided information from that image. If the images have already been tagged with labels of concrete objects, these are collected for further analysis. If they have not, an object recognition task especially attuned to art images is performed based on \cite{crowley2016art} in order to recognize physical objects in the image and gather labels for such objects. A similar approach is taken for extracting labels for depicted actions (e.g., “sitting”, “standing”) and/or relationships (e.g., “holding hands”, “hugging”). If tags for actions or relationships have already been attached to the images in the corpus, these are used. If not, we perform relation and action detection on the images following the most recent relation modeling techniques after they have been trained specifically to process art images, following the approach by \cite{kadish2021improving}. Finally, color analysis is performed on the images following the method provided by the extcolors Python package.\footnote{https://pypi.org/project/extcolors/} This technique groups colors based on visual similarities using the CIE76 formula, and outputs both an image (for visual representation) and a text result with the usage (in number and percentage of pixels) of the top five dominant (most used) colors in that image.

\section{Experiments and Results}
\label{experiments}

\begin{figure}[h]
  \centering
  \includegraphics[width=7cm]{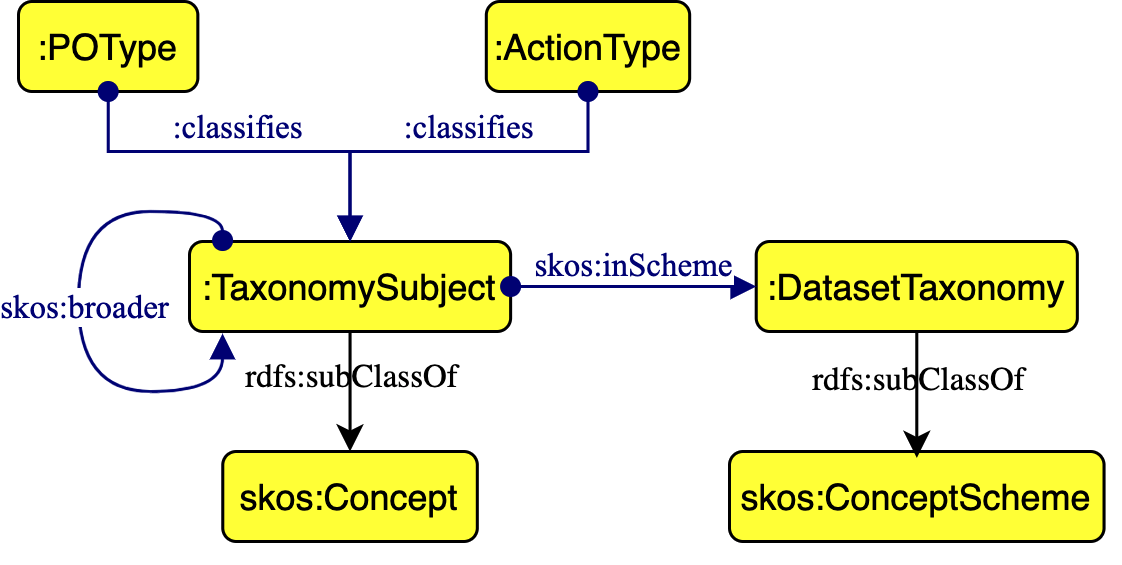}
  \caption{Addition to the MUSCO ontology model to formalize the use of concept schemes coming from collections' or other datasets' taxonomies, such as the Tate's subject taxonomy. All classes with no explicit namespace belong to the namespace of the MUSCO ontology.}
  \label{fig:T_BOx_addition}
\end{figure}

\noindent \textbf{Input Data Preprocessing.} In order to apply our method to Tate Gallery Collection dataset described in Section 2, it was first necessary to reconstruct the hierarchy of the Tate’s subject taxonomy in a way that allowed its eventual integration into a Social Concepts KG using the MUSCO ontology. For this task, we  performed a survey of the taxonomy based on metadata accessed in March of 2021, finding that (1) it has three hierarchical levels, going from broadest to narrowest categories, and that (2) it includes hierarchical relationships between 2409 subject tags. These 2409 subject tags were the ones used to reconstruct and then formalize the Tate’s tagging hierarchy. In order to integrate the taxonomy and its subjects into our KG, we extended the MUSCO ontology by reusing concepts defined in the Simple Knowledge Organization System (SKOS) data model\footnote{https://www.w3.org/TR/skos-reference/\#schemes} (see Figure \ref{fig:T_BOx_addition}). With this extension we were able to represent hierarchical relations between subject tags, specifically with the property \texttt{skos:broader}. We implemented a Python script to transform and serialise the taxonomy from JSON to Turtle (.ttl) format\footnote{https://github.com/tategallery/collection/tree/master/processed/subjects}. All resources (MUSCO ontology, .ttl file, Python functions) are available in the Github repository.\footnote{https://github.com/delfimpandiani/musco}


\medskip

\begin{figure}[h]
    \centering
    \includegraphics[width=14cm]{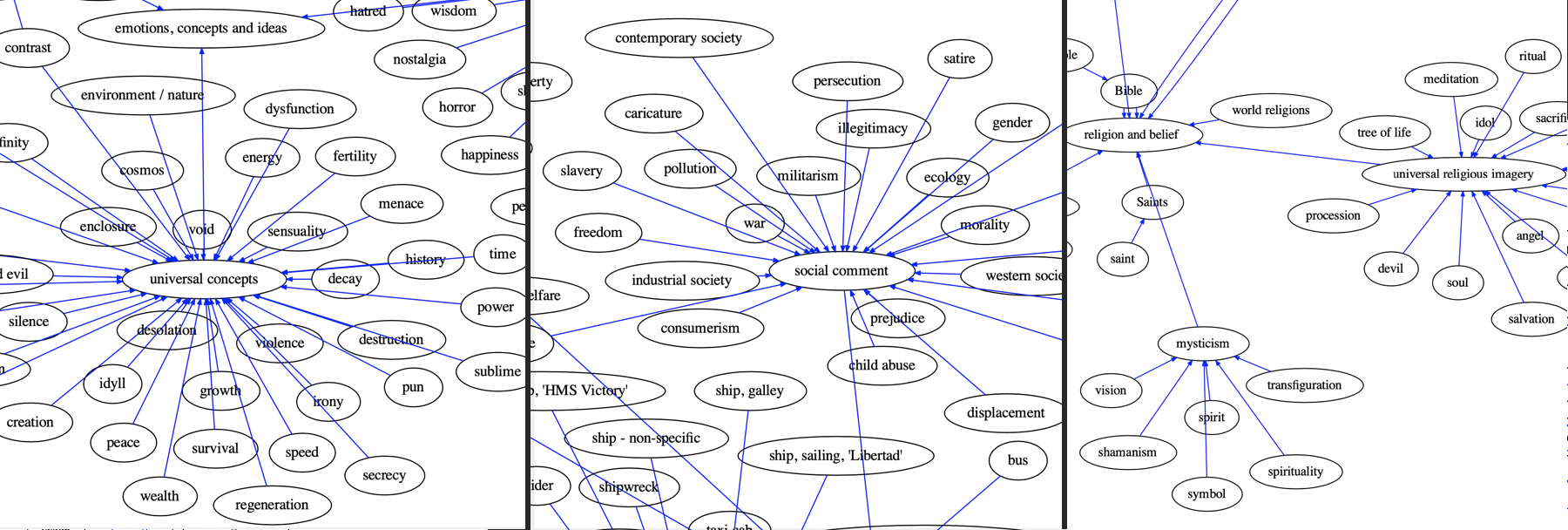}
    \caption{Three main areas of interest for the identification of social concepts within the Tate subject taxonomy. Social concepts such as “destruction”, “peace”, “wealth”, “courage” (surrounding  “emotions, concepts and ideas”), “consumerism”, “freedom”, “slavery”, “nationalism”, “ecology” (surrounding “society”), “magic”, “enchantment”, “worship”, “blessing” (surrounding “religion and belief”), among others, were identified in these areas.}
    \label{fig:areas}
\end{figure}

\noindent \textbf{Social Concepts Candidate List Creation.} Visualizations of the Tate’s subject taxonomy as graphs (also available in the GitHub repository) were performed with the Graphviz\footnote{https://pypi.org/project/graphviz/} package, in order to ease manual investigation of its coverage and identification of areas where social concepts may be more pervasive. Three areas of interest emerged: first, the level 0 category “emotions, concepts and ideas” (specifically its level 1 children “universal concepts” and “emotions and human qualities”); second, the level 1 category “social comment” (child of level 0 category “society”), and third, the level 0 category “religion and belief” (see Figure \ref{fig:areas}). A total of 166 “narrow” [level 2] social concepts were manually selected from these categories (80 from “emotions, concepts, and ideas”, 67 from “society”, and 19 from “religion and belief”). These concepts’ parent [level 1] and grandparent [level 0] tags were excluded from subsequent analysis for two reasons. First, compared to their narrow children tags (e.g., “fear”, “education”), many of the broader tags actually refer to multiple social concepts at once (e.g., “emotions, concepts and ideas”, “education, science and learning”). Secondly, a manual investigation of the Tate artworks’ metadata showed that artworks seem to be explicitly tagged with level 2 tags, and level 1 and level 0 tags are only included by virtue of being higher in the hierarchy. That is, there seem to be no Tate artworks that are tagged only with the broader level 1 or 0 tags. The complete list of social concepts and their parents is available in Appendix 1.

\medskip
\noindent \textbf{Concept-Artwork Matching.} For each of the 166 social concepts, the number of artworks tagged with that social concept was extracted and recorded. Additionally, data about each artwork, including its bibliographic information (name, id, artist, date, etc.) was stored.

\medskip

\noindent \textbf{Co-occurring Subject Tags.} For each social concept, an investigation of the metadata of the artworks tagged with it was performed, collecting other subject tags used to index the content of those artworks. In this way, for each social concept, we created a dictionary holding all co-occurring subject tags and the frequency of co-occurrence. This allowed for the collection of object and action tags without having to resort to object or human-object interaction recognition techniques.

\medskip

\noindent \textbf{Co-occurring Physical Objects.} Symbol grounding and \emph{acquired embodiment} is expected to occur with labels referring to physical referents. Therefore, we wanted to identify all co-occurring tags that specifically refer to physical objects. The Tate’s subject taxonomy allowed for a quick solution to this goal, as its hierarchical organization includes the top [level 0] concept ``objects'', which based on a manual investigation seem to refer to physical objects. We extracted all tags under this category. A manual inspection of the initial results showed that certain physical entities, such as ``woman'', or ``tree'' were not being extracted from artworks that clearly depicted these entities (and were tagged with them). From this observation, we noted that additional labels referring to physical objects were under the categories ``people'' and ``nature'', so we also extracted all tags whose parent tag was ``children'', ``adults'', or ``nude'', as well as those tags whose parent or grandparent tag was “nature”. While it was not possible to confirm that absolutely all terms referring to physical objects were extracted, after a manual inspection of the Tate's subject taxonomy, it was concluded that a large majority of them was indeed extracted. We then performed statistical analyses to obtain the number and frequency of co-occurring physical objects for each social concept, as well as the top ten most frequent physical objects co-occurring with each social concept. We also calculated the averages and medians of these measures by taking into account all social concepts and their co-occurring physical objects. 
\medskip

\noindent \textbf{Co-occurring Physical Actions.} A similar process to the one just described for labels referring to physical objects was performed for what, within the Tate’s subject taxonomy, are “action” labels. The identification of tags that refer to actions was completed by identifying tags under the categories “actions:postures and motions”, “actions:processes and functions”, or “actions:expressive”, and “animals:actions”. We extracted all action tags co-occurring with each of the social concepts, and then performed statistical analyses as described in the previous section.

\medskip

\noindent \textbf{Dominant Color Visual Analyses.} We then performed color analyses related to two social concepts selected as case studies: “consumerism” and “horror”. The hypothesis supporting this choice was that they should show distinctive color profiles. For each concept, the color analysis was performed with the extcolors Python package mentioned above, on 30 randomly selected images of artworks (specifically, of paintings and prints) that had been explicitly tagged with the concept. For each image, a color palette was created by extracting the RGB coordinates and occurrence rates (in both number and percentage of pixels) of the five colors with the highest occurrence rate in that image. The criterion for this choice was that identifying the colors with highest occurrence in an image is a proxy for identifying the most pervasive/visible colors in an image, which may be a relevant feature humans use to judge whether images evoke certain concepts. In order to represent this idea more intuitively, a further analysis was completed to generate images of the \emph{proportional} palettes, so as to represent the most common colors present in each of the artworks in a more intuitive way. This final task was completed with the MulticolorEngine by TinEye,\footnote{https://labs.tineye.com/color/} but we are developing code for the automation of this procedure.

\medskip

\noindent \textbf{Data Integration.} We are developing software to automatically incorporate the extracted data, including the co-occurrence patterns and visual features--into a Social Concepts KG.\footnote{The KG production is a work in progress that can be monitored on the GitHub repository.}

\subsection{Results}

Based on the metadata accessed in March of 2021, the Tate's subject taxonomy was found to be divided in three levels: [level 0] top concepts, representing the most general categories; [level 1] slightly narrower concepts, children of level 0 concepts; and [level 2] narrowest concepts, children of level 1 concepts and grandchildren of level 0 concepts. Out of the 2409 subject tags in the taxonomy, 16 are level 0 concepts, 142 are level 1 concepts, and 2251 are level 3 concepts. The 166 social concepts selected as initial candidates are all grandchildren of 3 out of the 16 top concepts, and children of 15 out of the 142 middle categories available (see Table \ref{tab:tax-levels}).

\noindent \textbf{Concept-Artwork Matching.} The number of Tate artworks explicitly tagged with each of the 166 chosen social concepts ranged from 368 (“death”) to 1 (“paranoia”), with an average of 48 matches and a median of 27 matches. The two case studies were in the top 20\% of social concepts ranked on artwork matches (“consumerism” with 71 artworks, and “horror” with 146 artworks) (see Table \ref{tab:matchstats}).

\medskip

\begin{figure}[h]
  \centering
  \includegraphics[width=1\textwidth]{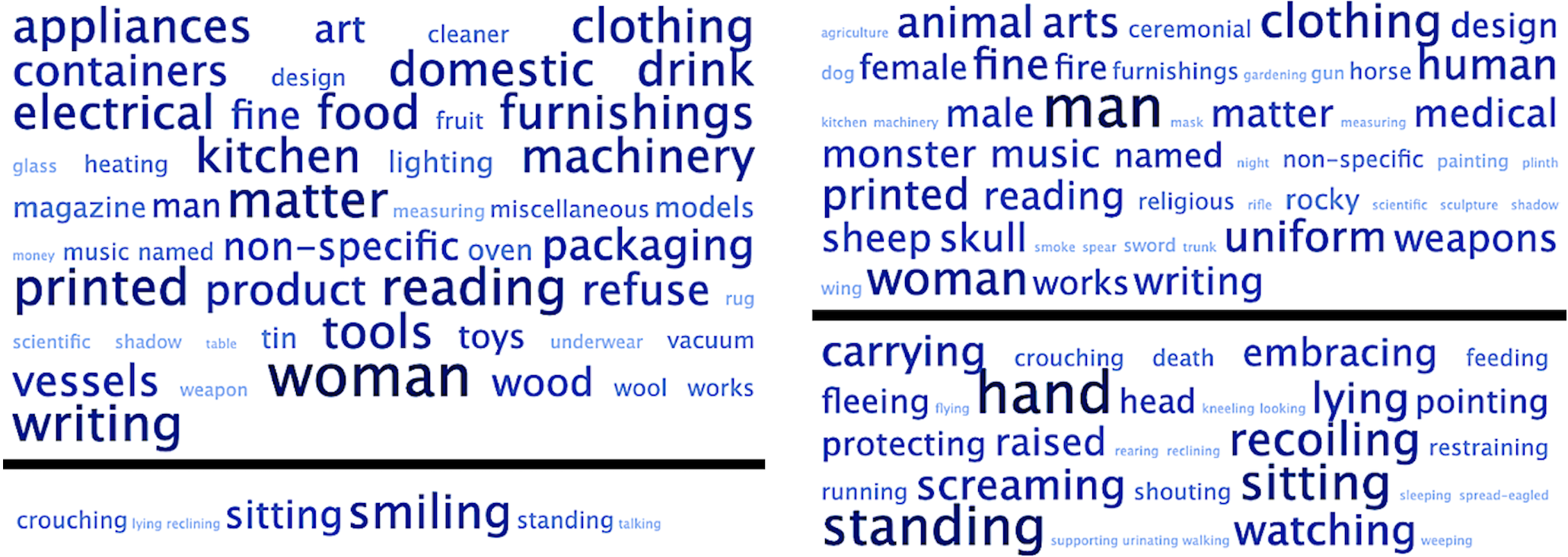}
  \caption{Wordclouds for the top 50 co-occurring object (top) and action (bottom) tags for all artworks tagged with ``consumerism'' (left) and with ``horror'' (right). Larger words more frequently co-occurred with the social concept of interest.}
  \label{fig:wordclouds}
\end{figure}

\noindent \textbf{Co-occurring Physical Objects and Actions.} The number of co-occurring tags for each of the 166 chosen social concepts ranged widely, from 1506 (“death”) to 7 (“paranoia”), with an average of 311 co-occurring subjects and a median of 262. Further analyses were performed to identify physical objects and action tags from the co-occurring subject tags for each social concept. The number of co-occurring \emph{physical object} tags for each of the chosen social concepts ranged from 288 (“death”) to 6 (“infinity”), with an average of 69 co-occurring physical objects and a median of 55 physical objects. Table \ref{tab:topcon} shows the top ten most frequent co-occurring physical objects for four social concepts. The number of co-occurring \emph{action} tags for the 166 social concepts was decisively smaller, ranging from 38 (“death”) to 0 (“void”), with an average of 11 co-occurring actions and a median of 8 actions. Table \ref{tab:topact} shows the top ten most frequent co-occurring actions for four social concepts. The average frequencies of co-occurrence for physical objects and for actions with each social concept are also presented in Table \ref{tab:matchstats}. Finally, Figure \ref{fig:wordclouds} includes more intuitive visual representations for most co-occurring physical objects (top) and actions (bottom) for the two case study social concepts.

\begin{table}[h]
\centering
\resizebox{\textwidth}{!}{%
\begin{tabular}{|l|l|l|l|l|}
\hline
\textbf{Level} & \textbf{\# Tags} & \textbf{Concrete Tags} & \textbf{\# Soc. Concepts} & \textbf{Soc. Concept Tags} \\ \hline
0 & 16 & “objects”, “nature”, “people” & 3 & “society” \\ \hline
1 & 142 & “weapons”, “trees”, “adults” & 15 & “social comment” \\ \hline
2 & 2251 & “missile”, “oak”, “old man” & 166 & “consumerism” \\ \hline
\end{tabular}%
}
\caption{Distribution of Tate subject tags based on its three hierarchical levels, going from broadest (0) to narrowest (2). Each level includes concrete concepts referring to physical objects and social concepts referring to non-physical objects.}
\label{tab:tax-levels}
\end{table}

\vspace{-2em} 

\begin{table}[h]
\centering
\resizebox{\textwidth}{!}{%
\begin{tabular}{|l|l|l|l|l|l|}
\hline
\textbf{Social Concept} & \textbf{\# Matches} & \textbf{\# CO Objects} & \textbf{Freq Top Objects} & \textbf{\# CO Actions} & \textbf{Freq Top Actions} \\ \hline
death & 368 & 288 & 71.1 & 38 & 14.5 \\ \hline
\textbf{horror} & 146 & 138 & 33.5 & 30 & 6.0 \\ \hline
\textbf{consumerism} & 71 & 129 & 16.4 & 7 & 2.4 \\ \hline
paranoia & 1 & 15 & 1.7 & 5 & 1.2 \\ \hline
\textit{Average} & \textit{48} & \textit{69} & \textit{9.7} & \textit{11} & \textit{2.8} \\ \hline
\textit{Median} & \textit{27} & \textit{55} & \textit{5.3} & \textit{8} & \textit{1.6} \\ \hline
\end{tabular}%
}
\caption{Number of artwork matches, co-occurring physical objects, and co-occurring actions, along with the average frequency of co-occurrence for the top ten most frequently co-occurring physical objects and top ten actions for four social concepts: ``death'' (with the highest number of matches), ``paranoia'' (lowest number of matches), ``consumerism'' and ``horror'' (case studies). Average and median values calculated from all  166 social concepts are also provided.}
\label{tab:matchstats}
\end{table}

\begin{table}[h]
\centering
\resizebox{\textwidth}{!}{%
\begin{tabular}{|l|p{13cm}|}
\hline
\textbf{Social Concept} & \textbf{Top 10 Physical Objects} \\ \hline
death & {`man', `woman', `religious and ceremonial', `clothing', `furnishings', `male', `weapons', `female', `fine arts and music', `sea'} \\ \hline
\textbf{horror} & {`man', `clothing', `woman', `uniform', `animal/human', `reading, writing, printed matter', `male', `fine arts and music', `monster', `medical'} \\ \hline
\textbf{consumerism} & {`woman', `reading, writing, printed matter', `clothing', `furnishings', `food and drink', `domestic', `electrical appliances', `kitchen', `tools and machinery', `product packaging'} \\ \hline
paranoia & {`man', `clothing', `woman', `figure', `male', `furnishings', `curtain', `jacket', `jumper', `suit'} \\ \hline
\end{tabular}%
}
\caption{Top ten most frequent co-occurring physical objects with the social concept with the most matched artworks (``death''), for the social concept with the least matched artworks (``paranoia''), and for the two case studies (``consumerism'' and ``horror'').}
\label{tab:topcon}
\end{table}

\begin{table}[h]
\centering
\resizebox{\textwidth}{!}{%
\begin{tabular}{|l|p{13cm}|}
\hline
\textbf{Social Concept} & \textbf{Top 10 Actions} \\ \hline
death & {`standing', `lying down', `reclining', `supporting', `embracing', `sitting', `flying', `kneeling', `carrying', `sleeping'} \\ \hline
\textbf{horror} & {`standing', `sitting', `recoiling', `watching', `lying down', `screaming', `carrying', `hand/hands raised', `embracing', `fleeing'} \\ \hline
\textbf{consumerism} & {`smiling', `sitting', `crouching', `standing', `reclining', `lying down', `talking'} \\ \hline
paranoia & {`sitting', `crouching', `reclining', `standing', `walking'} \\ \hline
\end{tabular}%
}
\caption{Top ten most frequent co-occurring actions with the social concept with the most matched artworks (``death''), for the social concept with the least matched artworks (``paranoia''), and for the two case studies (``consumerism'' and ``horror'').}
\label{tab:topact}
\end{table}

\begin{figure}[!h]
  \centering
  \subfloat[Consumerism]{\includegraphics[width=0.48\textwidth]{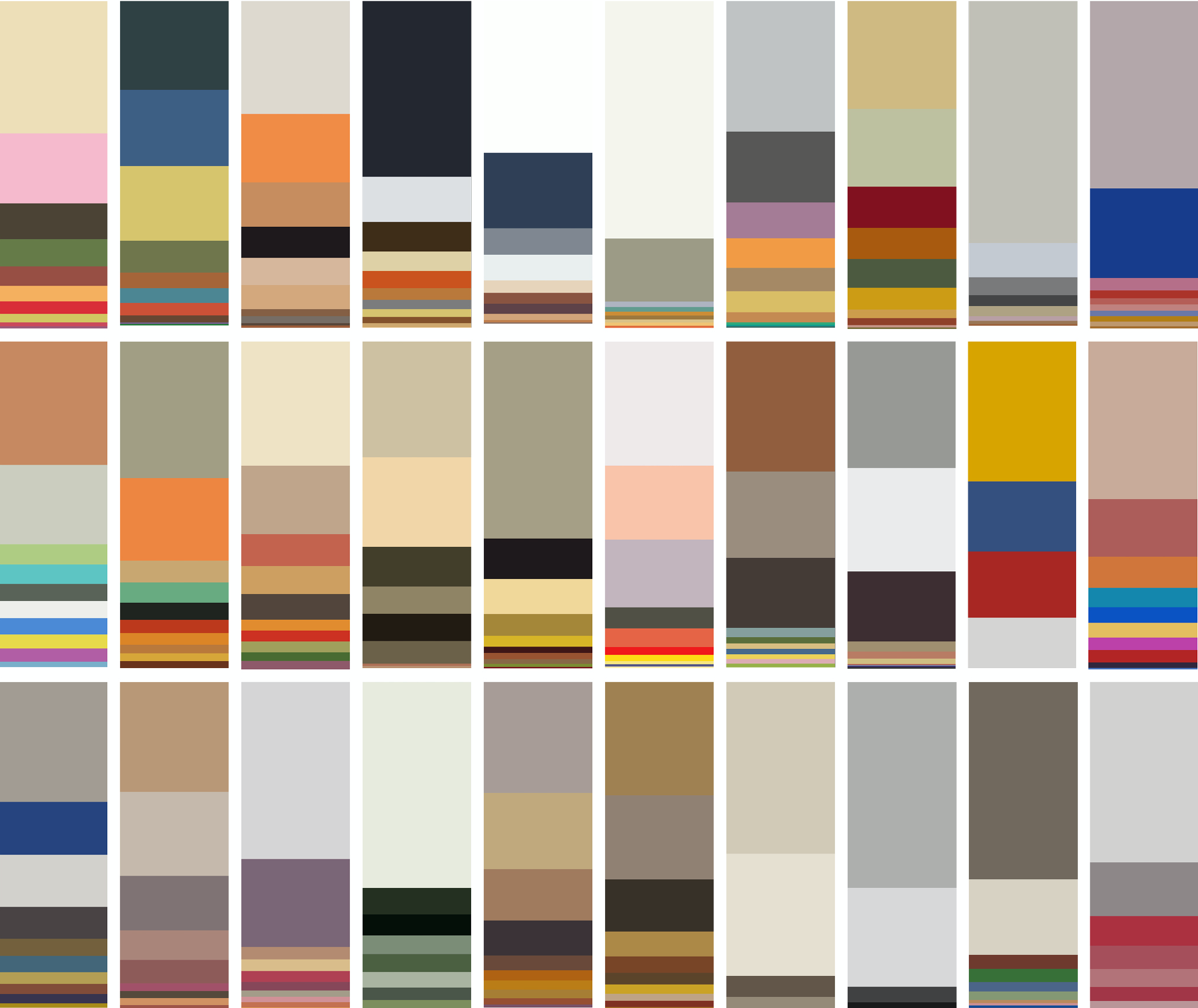}\label{fig:f1}}
  \hfill
  \subfloat[Horror]{\includegraphics[width=0.48\textwidth]{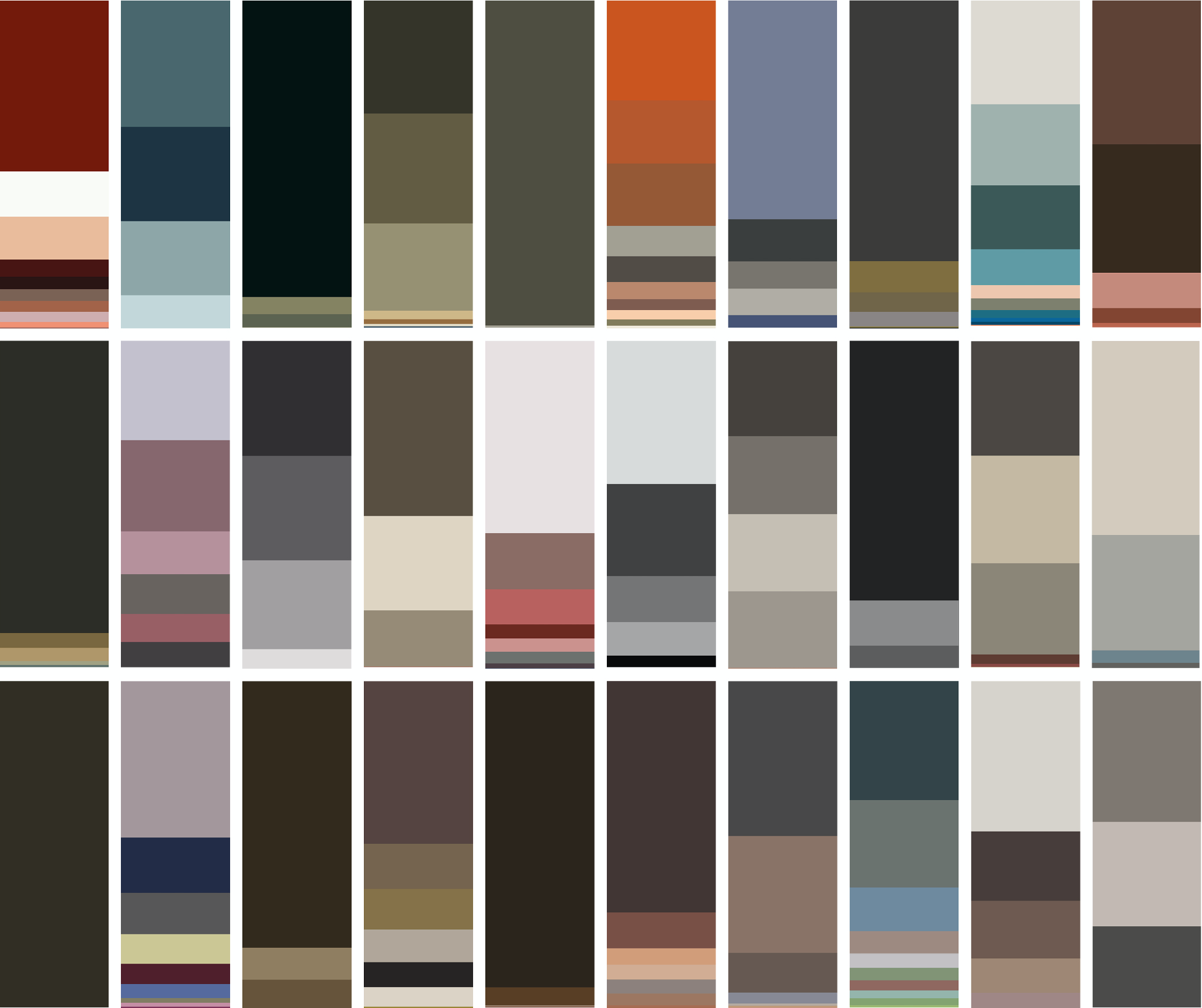}\label{fig:f2}}
  \caption{Proportional palettes of 30 paintings and print images tagged with ``consumerism'' (left) and with ``horror'' (right).}
\label{fig:palettes}
\end{figure}

\noindent \textbf{Dominant Color Visual Analyses.} Figure \ref{fig:palettes} presents visual representations (as proportional color palettes) of the color palette analyses performed on 30 randomly chosen images of paintings and prints for each of the two case study social concepts (“consumerism” and “horror”).



\section{Discussion}
\label{discussion}
In our reconstruction and examination of the Tate’s subject taxonomy, we found that social concepts referring to \emph{non-physical} objects were concentrated in three major areas of the taxonomy (“emotions, concepts, and ideas”, “society”, and “religion and belief”), categories which resemble cognitive subtypes of “abstract concepts”: emotional concepts, social concepts, and moral/ethical concepts. This suggests that our manual investigation may have covered a good amount of social concepts considered {\lq abstract\rq} in the cognitive sense. Gathered from the tagging taxonomy of a well-known collection of art images, the initial list of social concepts we present stands as empirical proof of the use of social concepts for tagging the content of visual material. With this work, we begin the creation of a corpus of (art) images tagged with social concepts, currently available in the GitHub repository as a dictionary with social concepts as keys, and lists of image urls as values. Additionally, our results show that the chosen social concepts were evoked by an average of 48 artworks, providing further evidence of the availability of data to pursue subsequent analyses.

Our experiments on the co-occurrence of physical objects and actions further show that it is possible to develop computational techniques that mirror the idea of symbol grounding and \emph{acquired embodiment}, i.e., it is possible to identify perceptual features of concrete objects and actions that co-occur with social concepts, at least in the context of art images. While there seems to be some regularity in the results, the low frequencies of co-occurrence suggest that further research is needed to understand which of these objects and actions, if any, have a substantial effect on the evocation of a social concept. A manual examination of the top ten most frequent concepts co-occurring with social concepts (some of them presented in Table \ref{tab:topcon} and Table \ref{tab:topact}) suggest some regularity in the physical objects and actions that most frequently co-occur with certain social concepts in these art images. Most of these co-occurrences seem to agree with intuition (i.e., “consumerism” co-occurring with “clothing”, “food and drink”, and “product packaging”; “horror” co-occurring with “monster”, “recoiling” and “screaming”). 
The results of the color analyses, visible in the proportional color palettes presented in Figure \ref{fig:palettes}, also strongly suggest a certain degree of regularity in the color features of Tate art images that evoke certain social concepts. As with the top co-occurrence results, at first sight the color palettes seem to agree with intuition (i.e, “consumerism” showing a greater luminosity and variety of bright color, as in the aisles of a supermarket; “horror” showing dark and less varied colors and tones, as in typical scenes from horror movies).

Some limitations of this work should be noted, such as the fact that the images used in the experiments were not distinguished by type of artwork medium (painting, print, drawing), which may have an effect on the type and frequency of features that were extracted from the images. It is also important to note that the regularity observed from the experimental results may be limited only to the Tate’s collection and not generalisable to other art image catalogues or collections. That is, while the Tate dataset provided us with a clean, annotated corpus in which social concepts are explicitly used as tags, as a curated dataset it encompasses a limited geographic, historical and cultural perspective. Additionally, the fact that it is expert-tagged might result in a biased interpretation and classification of the artworks, which might differ from the interpretations of a larger, more diverse group of viewers. Given social concepts' ambiguous meanings, further work is needed which takes into account  various possible interpretations of the same artwork by  different observers.

\section{Conclusion and Further Directions}
\label{conclusion}
 

Our approach models social concepts as complex objects, which can be described with multimodal frames that integrate multisensory information. To represent this model formally, we designed the pattern-based ontology Multimodal Descriptions of Social Concepts (MUSCO), which allows for the conjunct semantic characterization of multimodal features. A crucial next step is to analyse the suitability of the proposed ontology for practical usage on datasets from different sources. Our approach also proposes a pipeline for automatically extracting and integrating features of images that evoke social concepts. We show its potential by testing it on a corpora of art images from a well-known collection. Our experimental results point towards some regularity in certain sensory features of images tagged with specific social concepts, and open space for further research to evaluate the proposed approach focusing on datasets with different characteristics. More than anything, our results serve as a proof of concept and open up new lines of future research. The automatic population of a Knowledge Graph (KG) with the extracted data is the natural next step of this work, potentially through mapping-based knowledge extraction techniques.
The generated KG may be additionally populated with other methods, for example by automatically generating descriptive paragraphs from the art images, and then performing knowledge extraction on the natural language paragraphs, as well as by including additional sensory data, such as sound or smell data, that may evoke certain social concepts. With this work, we also begin creating a corpus of (art) images that evoke certain social concepts. Moving forward, it ought to be enlarged and diversified after a survey of additional catalogues and collections.

To further our overall goal, we can improve our approach by refining the initial social concepts list through alignment with the latest cognitive science research, as well as through user-based studies. Additionally, disambiguating the terms, expanding the terminology by leveraging lexical resources such as WordNet, VerbNet, and FrameNet, and studying the terms' distributional linguistic features in a textual corpus are next steps that we expect will lead to substantial improvements. The visual analyses can be further refined by distinguishing artwork medium types, and by extracting contrast measures, common shapes, repetition, and other visual patterns. A particularly interesting research direction would be to include facial recognition analyses in our pipeline, which may allow identification of emotions expressed by depicted subjects.

Our method is impulsed by the intuition that a KG containing multimodal data for social concept description can eventually serve as input to a learning model to automatically detect social concepts in images. 
Overall, the experiments performed and results presented in this paper serve as a proof of concept that extracting, integrating, and coincidentally exploiting multimodal data related to social concepts is a promising direction for future research. 


\bibliography{biblio}

\end{document}